\documentclass[letterpaper, 10 pt, conference]{ieeeconf}  

\usepackage[markup=bfit, deletedmarkup=sout, authormarkup=superscript]{changes}
\usepackage{amsfonts}

\usepackage{caption}
\usepackage{subcaption}
\usepackage{amsmath, xparse}
\usepackage{cleveref}
\usepackage{overpic}
\usepackage{graphicx}
\usepackage[export]{adjustbox}

\IEEEoverridecommandlockouts                              

\title{\LARGE \bf
 Volumetric Data Fusion of External Depth and Onboard Proximity Data For Occluded Space Reduction
}

\author{Matthew Strong$^{*}$, Caleb Escobedo$^{*}$, and Alessandro Roncone 

\thanks{$^{*}$Matthew Strong and Caleb Escobedo contributed equally to this work.
All authors are with the Department of Computer Science, University of Colorado Boulder, 1111 Engineering Drive, Boulder, CO USA {\tt\small name.surname@colorado.edu}}%

}

\begin{document}

\maketitle
\thispagestyle{empty} 

\ieeefootline{\small{\emph{4\textsuperscript{th} Workshop on Proximity Perception}\\ 2021 IEEE/RSJ International Conference on Intelligent Robots and Systems (IROS 2021), Prague, Czech Republic\\
Published under the Creative Commons license (CC BY-NC-ND) at KITopen}}

\begin{abstract}
In this work, we present a method for a probabilistic fusion of external depth and onboard proximity data to form a volumetric $3$-D map of a robot's environment. We extend the Octomap framework to update a representation of the area around the robot, dependent on each sensor's optimal range of operation. Areas otherwise occluded from an external view are sensed with onboard sensors to construct a more comprehensive map of a robot's nearby space. Our simulated results show that a more accurate map with less occlusions can be generated by fusing external depth and onboard proximity data.
\end{abstract}

\section{INTRODUCTION}

Roboticists have begun to investigate robot capabilities in cluttered environments where obstacles can be occluded from view of external depth sensors. In order for robots to plan safe trajectories, the robot motion planner requires an accurate spatial depiction of the robot's workspace. Traditionally, object positions surrounding the robot are estimated using depth data from externally mounted cameras, such as the Microsoft Kinect. However, the depth data provided from externally mounted cameras is prone to occlusions when objects are placed between the camera and the robot itself, as noted in \cite{navarro2021proximity}. Occlusions lead to an incomplete view of the robot's workspace that make it difficult to plan safe paths. To gain information about areas otherwise occluded from external depth sensors, \cite{navarro2021proximity} highlighted the benefits of using distance sensors placed on the robot's body.

Previous work using onboard proximity sensors has focused on utilizing information immediately for contact avoidance or anticipation, as shown in \cite{ding2021improving, escobedo2021contact}. External depth cameras have also been utilized for contact avoidance \cite{flacco2012depth}, and are frequently used to map a robot's surrounding space. To add additionally utility to data measured by onboard proximity sensors, in this paper, we fuse external depth and onboard proximity data to create a volumetric representation of the robot's environment. We extend the Octomap \cite{hornung2013octomap} framework to include onboard proximity sensor units along with external depth data. Our results demonstrate that otherwise occluded areas can be mapped using onboard proximity sensors. 

\section{Multi-modal Octomap Generation}

In our simulated environment, a depth camera is placed to give a third person view of the robot's workspace, and proximity sensors are placed on the robot's end-effector. We then process the information obtained from each sensor to compute the position of objects near the robot. Then, extending the Octomap framework, we modify our map using a novel probabilistic update, dependent on which sensor detected each object.

\subsection{Object Detection}
\subsubsection{Onboard Proximity Sensors} We position each proximity sensor so that its distance measurement is aligned with the sensor's $z$-axis. Accordingly, the position of an object $\mathbf{h_k} \in \mathbb{R}^3$ detected by sensor unit $\mathbf{k}$ can be computed as: 
\begin{equation}
     \mathbf{h_k} = {}^{O}\vec{r}_{PS_k} + {}^{O}{R}_{PS_k}\begin{bmatrix}
0 & 0 & d_{obs, k}
\end{bmatrix}^T ,
\label{eq:su_object_position}
\end{equation}
where $PS_k$ is the $k$th proximity sensor, and ${}^{O}\vec{r}_{PS_k} \in \mathbb{R}^3$ is the position of $PS_k$ with respect to the robot's base frame $O$. ${}^{O}{R}_{PS_k} \in \mathbb{R}^{3 \times 3}$ is the rotational matrix of $PS_k$ with respect to the robot's base frame. $d_{obs, k}$ is the sensed distance from the sensor to the object.
\subsubsection{Depth Camera}The computation of an object's position for a depth camera is given by:
\begin{equation}
     \mathbf{h} = {}^{O}\vec{r}_{DC} + {}^{O}{R}_{DC}\begin{bmatrix}
p_x & p_y & p_z
\end{bmatrix}^T ,
\label{eq:kinect_object_position}
\end{equation}
where $DC$ is the depth camera, and ${}^{O}\vec{r}_{DC} \in \mathbb{R}^3$ is the position of $DC$ with respect to robot's base frame. ${}^{O}{R}_{DC} \in \mathbb{R}^{3 \times 3}$ is the rotational matrix of $DC$ with respect to the robot's base frame. $p_x, p_y, p_z$ composes the $3$-D position of a detected point in the sensor's frame.

\subsection{Volumetric Mapping}
\label{sec:volumetric_mapping}

Octomap is an occupancy grid mapping framework based on octrees, which generates a probabilistic occupancy estimation given noisy depth data \cite{hornung2013octomap}. It explicitly calculates occupied and free space probabilities. As sensor measurements are read, they are used to update the occupation probability of nodes within the octree. $P(n | z_{1:t})$, the probability of node $n$ being occupied conditioned on $z_{1:t}$, depends on the current measurement $z_{t}$, a prior probability $P(n)$, and the previous probability estimate $P(n | z_{1:t-1})$. We refer interested readers to Equation 1 in \cite{hornung2013octomap} for more information. To reduce computation time, Octomap uses the log-odds of $P(n)$ to update a node's occupation probability. The log-odds is expressed as: 
\begin{equation}
\mathrm{L}(n)=\log \left[\frac{P(n)}{1-P(n)}\right]
\end{equation}
The log-odds of node $n$ being occupied conditioned on sensors' measurements from time $1$ to $t$ can be expressed as:
\begin{equation}
    L(n | z_{1:t}) = L(n | z_{1:t-1}) + L(n | z_{t}),
\label{eq:normal_log_odds_update}
\end{equation}
    \begin{figure}
        \centering
        \begin{subfigure}[b]{0.23\textwidth}
            \centering
            \frame{\includegraphics[width=\textwidth,height=4cm]{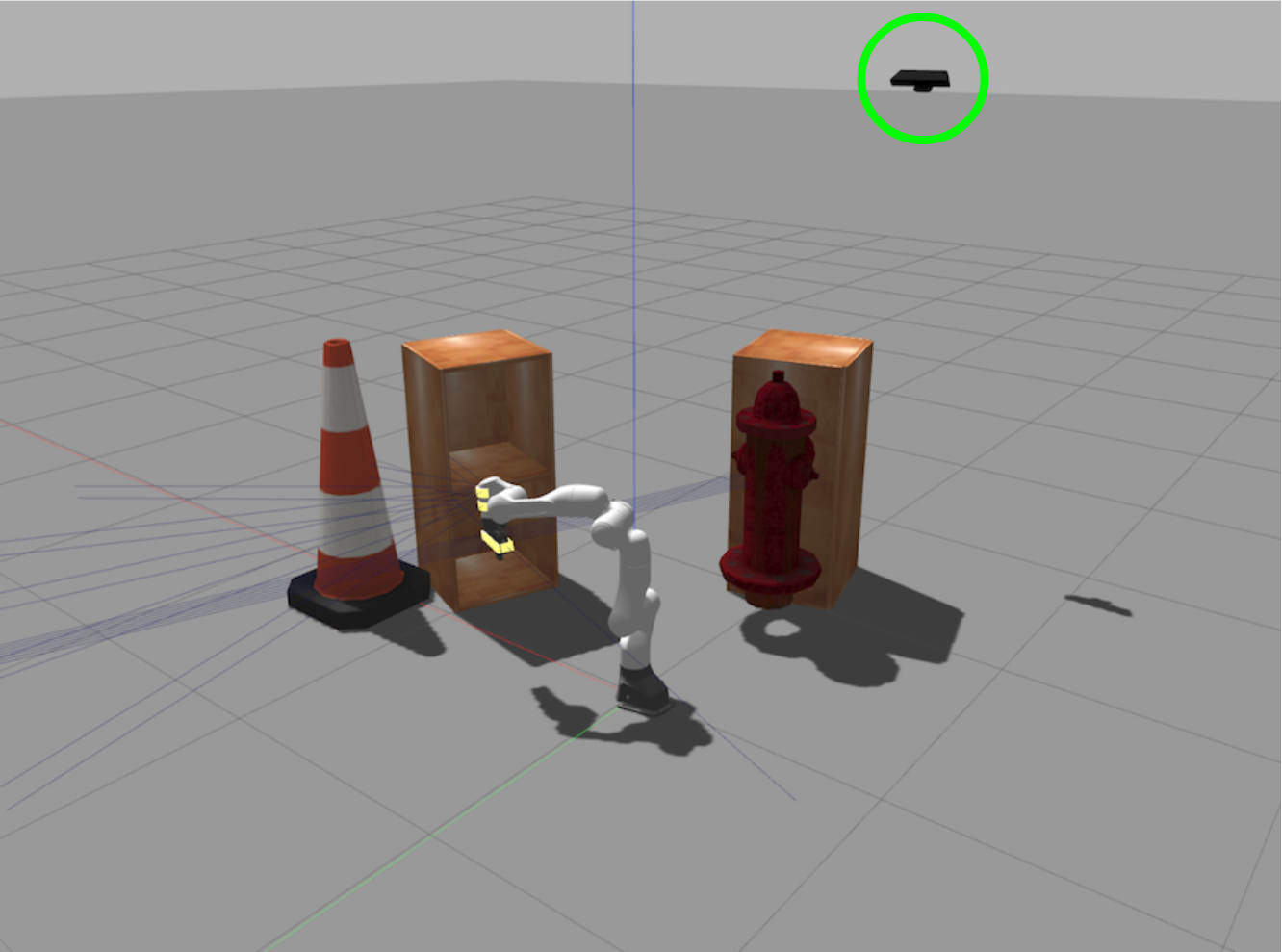}}
            \caption[Network2]%
            {{\small Gazebo Simulation}}    
            \label{fig:gazebo}
        \end{subfigure}
        \begin{subfigure}[b]{0.23\textwidth}   
            \centering
            \frame{\includegraphics[width=\textwidth,height=4cm]{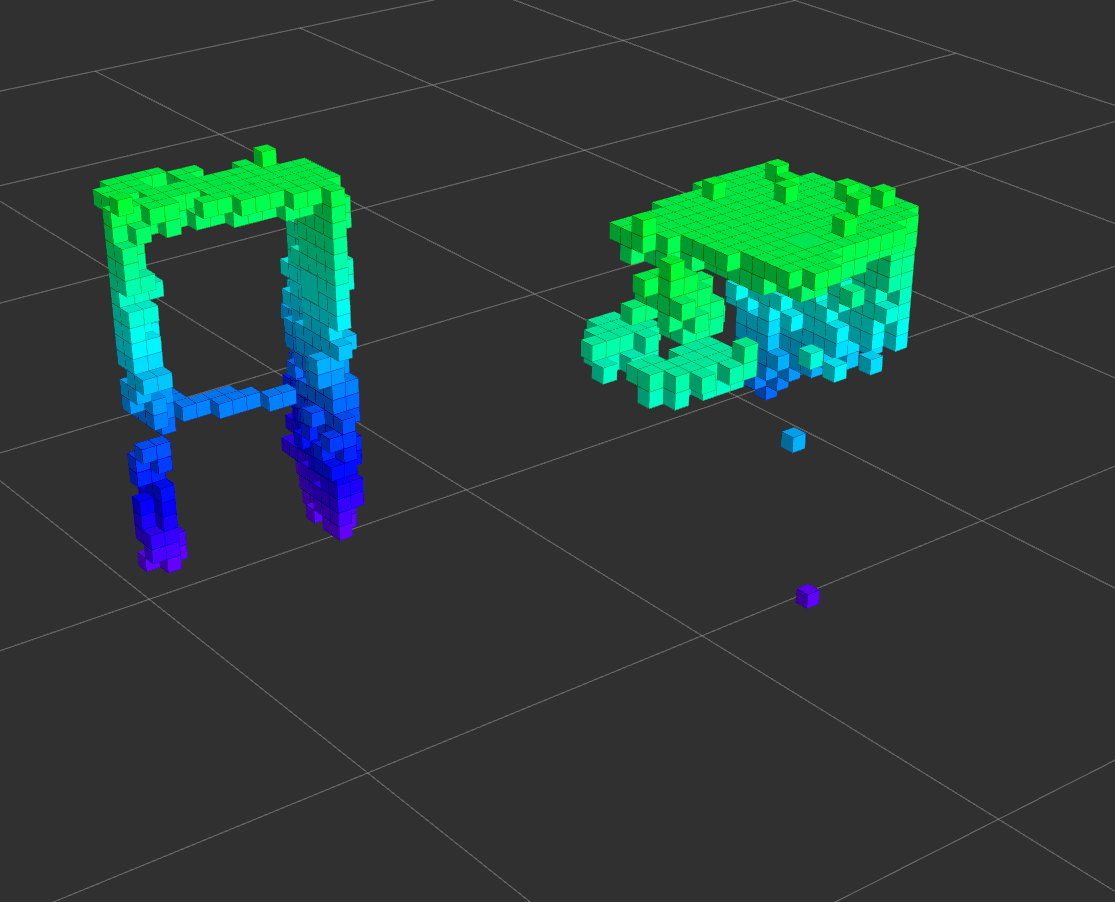}}
            \caption[]%
            {{\small Depth Camera}}    
            \label{fig:kinect}
        \end{subfigure}
        
        \begin{subfigure}[b]{0.23\textwidth}  
            \centering 
            \frame{\includegraphics[width=\textwidth,height=4cm]{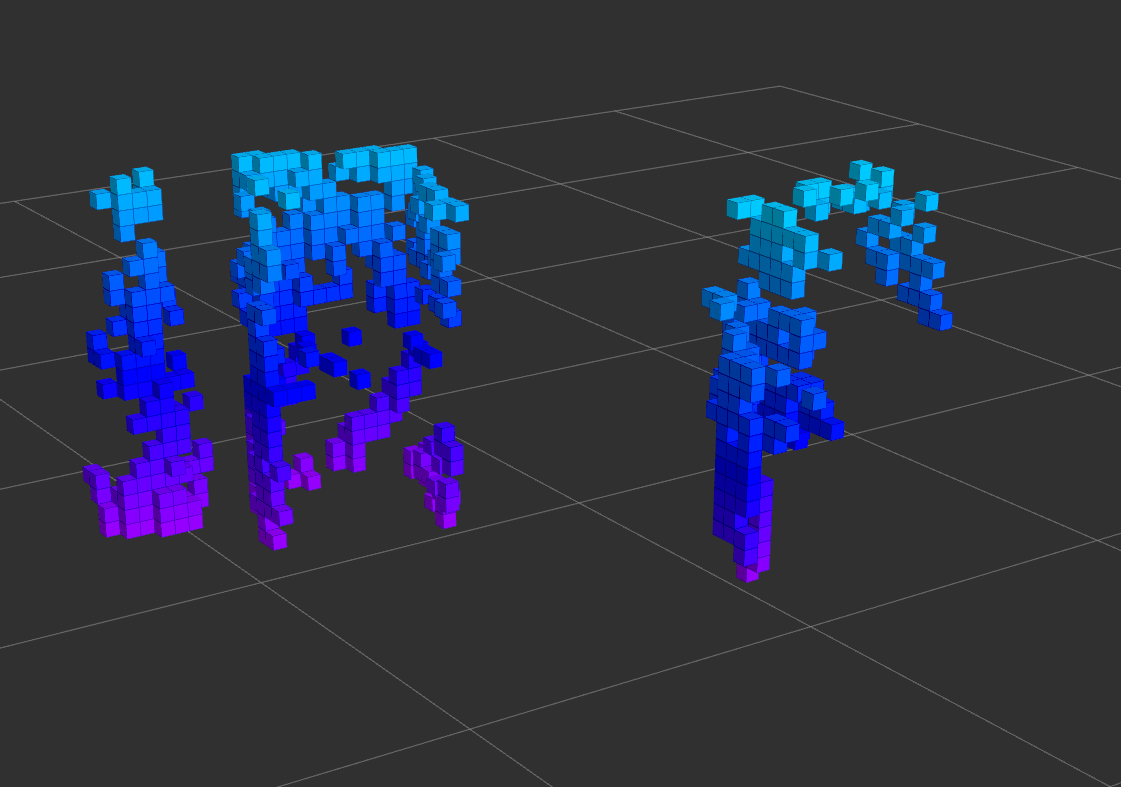}}
            \caption[]%
            {{\small Onboard Proximity}}    
            \label{fig:proximity}
        \end{subfigure}
        \begin{subfigure}[b]{0.23\textwidth}   
            \centering 
            \frame{\includegraphics[width=\textwidth,height=4cm]{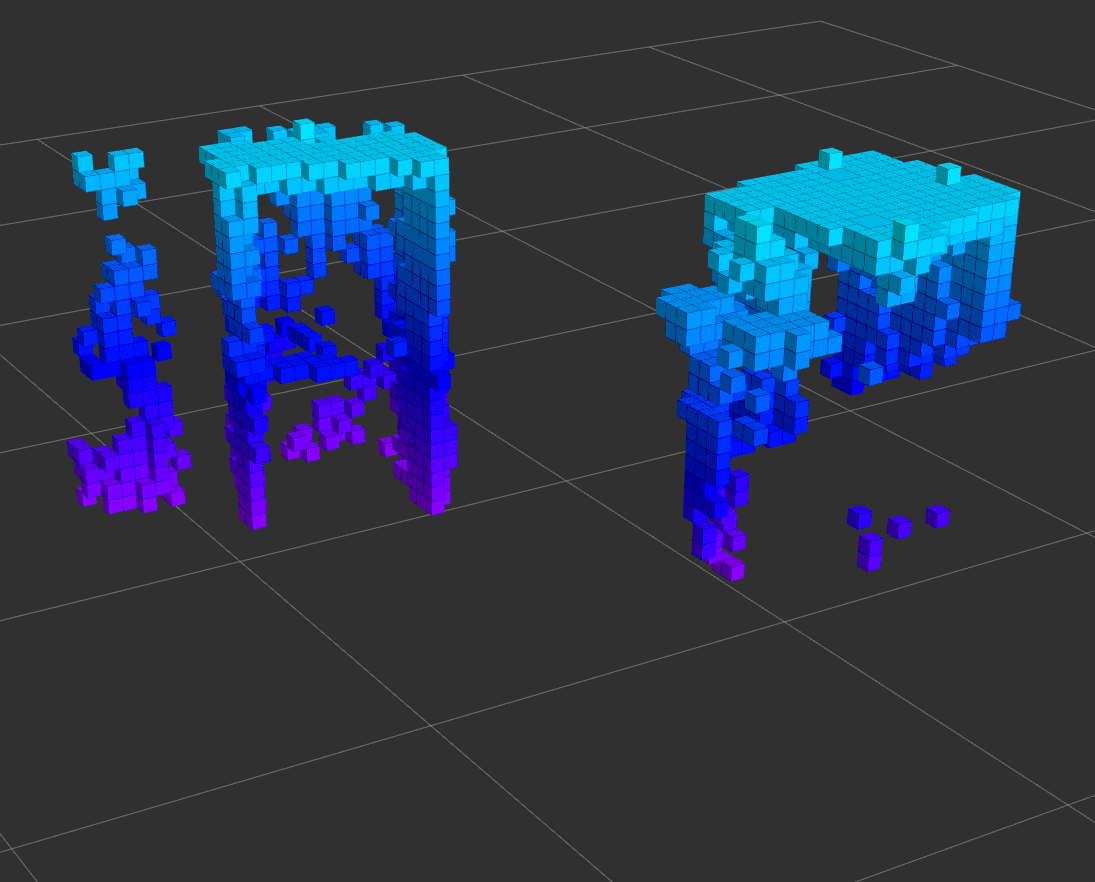}}
            \caption[]%
            {{\small Both Sensors}}    
            \label{fig:both}
        \end{subfigure}
        
        \begin{subfigure}[b]{0.23\textwidth}  
            \centering 
            \frame{\includegraphics[width=\textwidth,height=4cm]{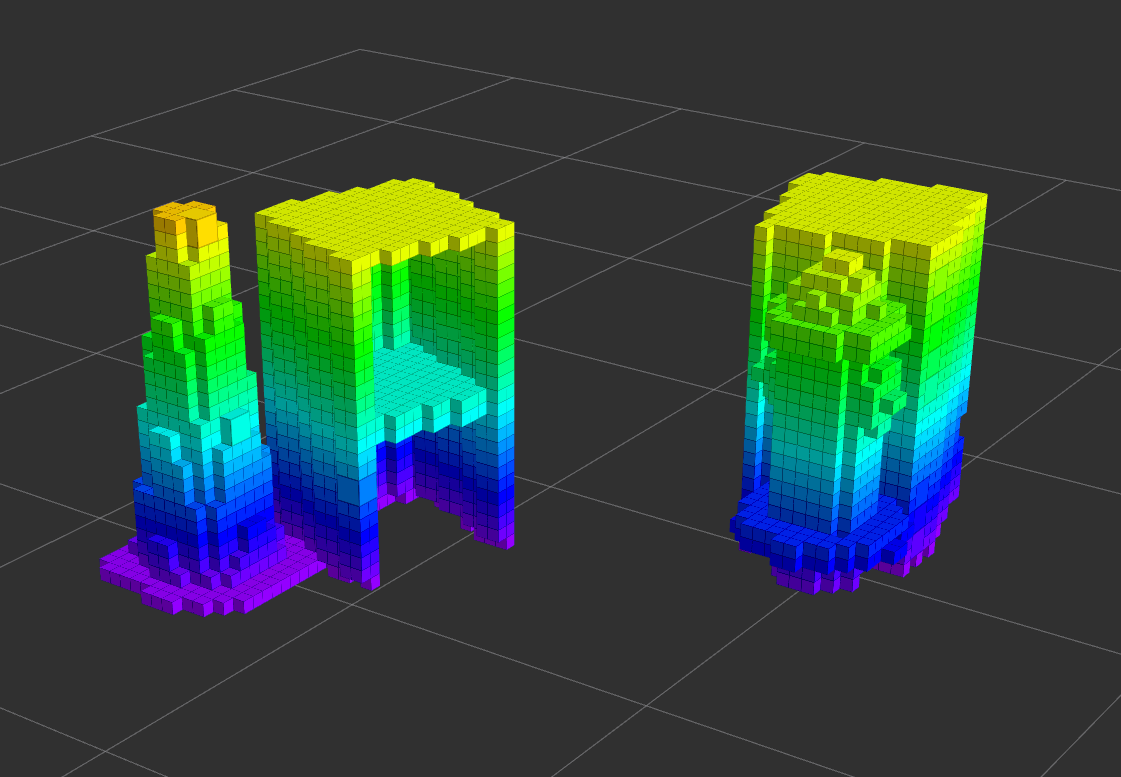}}
            \caption[]%
            {{\small Ground Truth Front}}    
            \label{fig:full_scene_front}
        \end{subfigure}
        \begin{subfigure}[b]{0.23\textwidth}   
            \centering 
            \frame{\includegraphics[width=\textwidth,height=4cm]{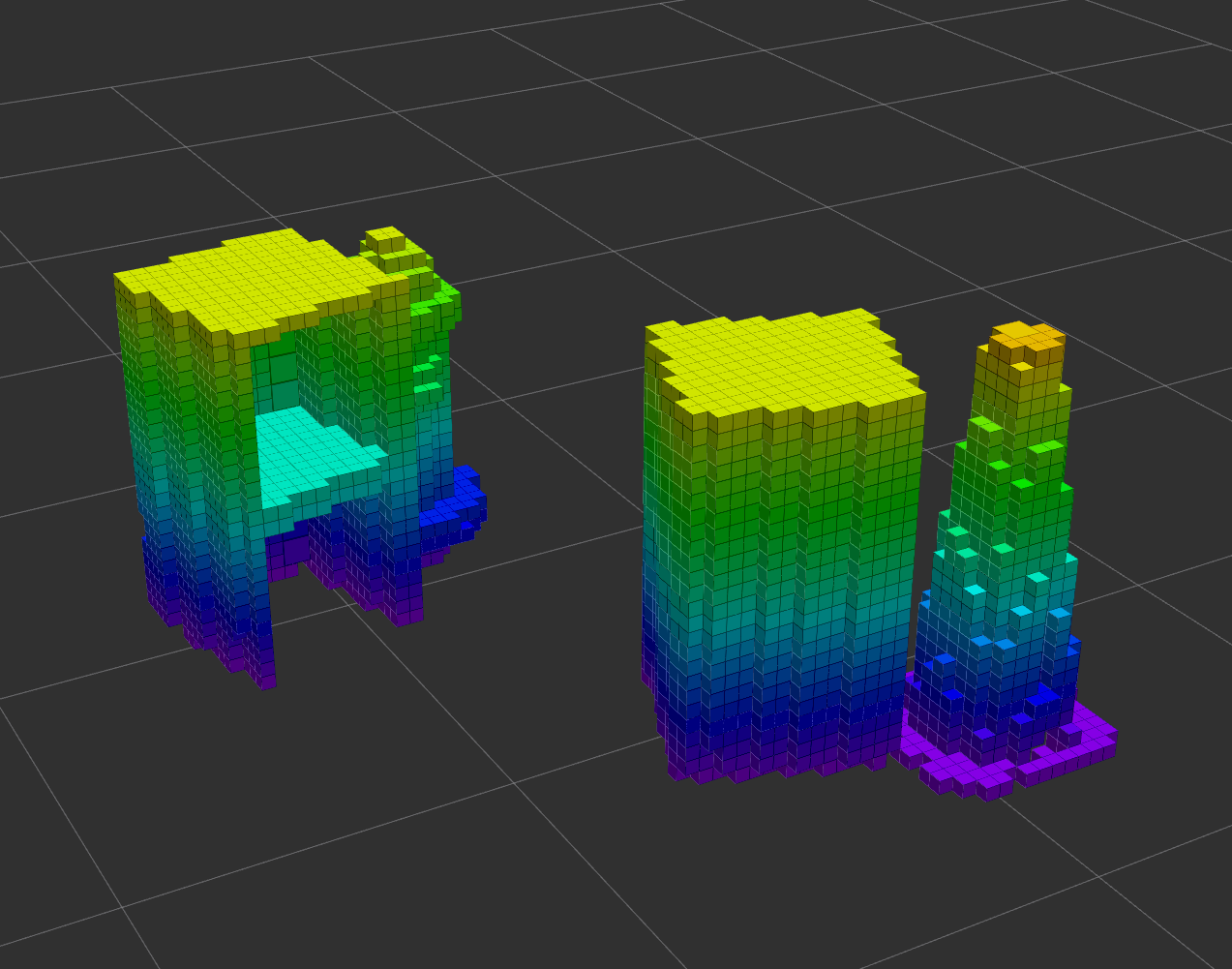}}
            \caption[]%
            {{\small Ground Truth Back}}    
            \label{fig:full_scene_back}
        \end{subfigure}
        \caption{Image (a) shows our simulated Gazebo environment with multiple objects occluded from the view of a depth camera, circled in green. Images (b), (c), and (d) show an Octomap representation when using specific sensor data. The full scene used to evaluate our method is shown in (e) and (f).} 
        \label{fig:map_comparison}\vspace{-16pt}
    \end{figure}
where $L(n | z_{1:t-1})$ is the prior log-odds and $L(n | z_t)$ is the log-odds of a node conditioned on the current sensor measurement \cite{hornung2013octomap}. Furthermore, the probability of a node being occupied based on the current sensor data is expressed in log-odds terms as:
\begin{equation}
    P(n | z_t) =   1 - \dfrac{1}{1 + \exp(L(n | z_t))},
\label{eq:log_odds}
\end{equation}

In our work, we modify the log-odds update notation to allow for multiple proximity sensors along with an external depth camera. Our log-odds update at time $t$ can then be written as:
\begin{equation}
   L(n | z_{t}) = L(n | DC_t) + \sum_{k=1}^{M} L(n | PS_{k, t}),
\label{eq:big_log_odds_update}
\end{equation}
where $M$ is the amount of onboard proximity sensors. $L(n | PS_{k, t})$ is the log-odds of node $n$ given data from proximity sensor $k$ at time $t$, and $L(n | DC_t)$ is the log-odds given data from depth camera $DC$. $L(n | PS_{k, t})$ is computed as:
\begin{gather}
        L(n | PS_{k, t}) = \begin{cases}
        0:  {d_{PS_{k, t}}} < 0.04, \\
         -0.07d_{PS_{k, t}} + 1 : {d_{PS_{k ,t}}} \geq 0.04, \\
         -0.4: \text{if beam traverses through node},
        \end{cases}
        \label{eq:log_odds_update} 
\end{gather}
where $d_{PS_{k, t}}$ is the distance from proximity sensor $k$ at time $t$ to a sensed object that is in node $n$. If a beam (a ray of points connecting the sensor to the detected object) passes through a node, which means that there is no detected object in that specific node, then a negative log-odds update is received (here, $-0.4$ corresponds to a probability of $0.4$ of node $n$ being occupied).
$L(n | DC_t)$ is:
\begin{gather}
        L(n | DC_t) = \begin{cases}
         0: {d_{DC_t}} \leq 0.5, \\
        -0.1d_{DC_t} + 1:  d_{DC_t} \ge 0.5, \\
        -0.4: \text{if beam traverses node},
        \end{cases}
        \label{eq:log_odds_update} 
\end{gather}
where $d_{DC_{t}}$ is the distance from the depth camera to the sensed object. 

For proximity data, we base our formulation on the official datasheet of the VL53L1X ToF sensor, which is the sensor we used in our previous real world experiments \cite{escobedo2021contact}. Its minimum range is $4$ cm. The ranging error for the VL53L1X ToF sensor in different test cases of varying distances is $2$ cm; the data sheet also shows that as the range of sensed data increases, so does the standard deviation of measurements \cite{st:VL53L1X}. With this information taken into account, we linearly decay the log-odds update value of proximity sensor data.

The depth camera log-odds value at a distance of $0.5$m and below is $0$ (which does not change the probability of a node being occupied) because the operational range of the depth camera starts from $0.5$m \cite{tolgyessy2021evaluation}. 
At distances greater than $0.5$m, it has been shown that the ranging error (accuracy and precision) of ToF (Time-of-Flight) depth cameras slowly increases over time; hence we apply a linear decay on the log-odds value of a detected node from the depth camera \cite{he2017depth}. We model our depth camera based on the Kinect-V2; its error is $3$cm at $2$m \cite{he2017depth}.

\begin{figure}[h]
\centering
\includegraphics[width=0.5\textwidth]{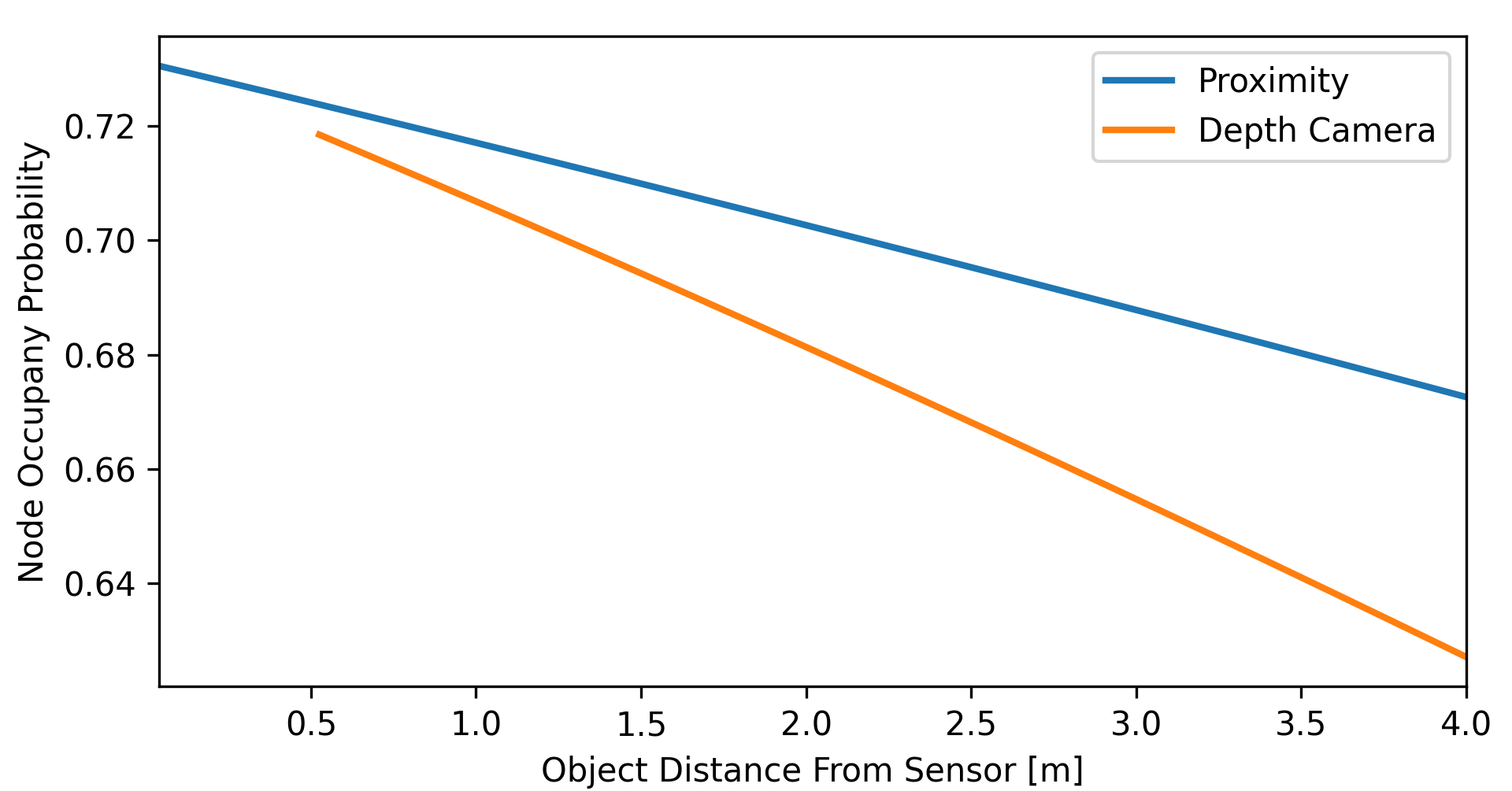}
\caption{Probabilities of a node being occupied for proximity sensor and depth camera measurements based on distance, computed from our log-odds update.}
\label{fig:update_graph}
\vspace{-6pt}
\end{figure}

We graph the corresponding probabilities from our log-odds updates in \cref{fig:update_graph}.
The probability of a node being occupied with the depth camera from $0.5$m to $4$m decreases from $0.73$ to $0.68$, and the occupation probability with a proximity sensor from $0.04$m to $4$m decreases from $0.72$ to $0.64$. These values are centered around the optimal constant update value introduced in the original Octomap paper \cite{hornung2013octomap}. These values were set based on both sensor's respective data sheets and research analysis of performance from \cite{he2017depth, tolgyessy2021evaluation}. 
\section{Results}
\subsubsection{Setup}
To evaluate our method, we conduct experiments in the Gazebo simulator using C++ and ROS with a simulated, 7-DoF Franka Emika Panda robotic manipulator. We simulate $34$ proximity sensors, placed on the robot end-effector, and an externally mounted depth camera. We publish batched proximity data at a rate of $30$ Hz with $2$ cm noise, while the depth camera data is published at a rate of $10$ Hz with $3$ cm noise. Noise is based on sensor error introduced in \cref{sec:volumetric_mapping}.
Our results compare the amount of free and occupied space generated from proximity sensor data, depth camera data, and both data streams fused, as displayed in \cref{fig:map_comparison} with a ground truth map. Our simulated environment is shown in \cref{fig:gazebo}. 
During simulation, the robot moves in a circular path with a radius of $0.3$ m at a speed of $0.188$ m/s.
\begin{table}[h!]
\centering
\begin{tabular}{||c | c | c | c | c ||} 
 \hline
  Sensors Used & Occupied & Free & Missed & Incorrect\\%
 \hline\hline
 Depth Camera & 970 & 27510 & 3523 & 299\\ %
 \hline
 Proximity & 587 & 47847 & 3697 & 26\\%
 \hline
 Proximity + Depth & \textbf{1430} & 66697 & \textbf{2469} & 317\\ %
 \hline
\end{tabular}
\caption{This table shows a comparison of our generate maps from a specific sensor type to a ground truth map shown in \cref{fig:full_scene_front}. \textbf{Occupied}: A node in space is occupied in both ground truth and generated map. \textbf{Free}: Node is free in both ground truth and generated map. \textbf{Missed}: Node is occupied in ground truth but does not exist in the generated map. \textbf{Incorrect}: Node is not occupied in ground truth, but is occupied in the generated map.}
\label{table:results}\vspace{-6pt}
\end{table}
\subsection{Volumetric Comparison}

Our simulated environment can be seen in \cref{fig:gazebo} with two objects purposefully occluded by large shelves placed between the robot and the camera. The cone in \cref{fig:gazebo} is only sensed by the proximity sensors, as seen in \cref{fig:proximity}. 
The onboard proximity sensors' information supplements the depth camera data to give a representation of the shelf as two distinct open compartments. 

We compare our maps generated shown in \cref{fig:map_comparison} to a ground truth mapping created with a manually controlled depth camera with no noise as shown in \cref{fig:full_scene_front} and \cref{fig:full_scene_back}.
In our evaluation, we compare the ground truth octree with a constructed octree, both of which have the same minimum octree resolution of $0.04$m.
From \cref{table:results}, the proximity and depth map has $1430$ occupied nodes that matched the ground truth, as compared to $970$ for the depth camera and $587$ for proximity. The amount of correctly free nodes is also significantly higher for the proximity and depth map. Our method reduces the amount of missed occupied cells, while only slightly increasing the amount of incorrect nodes.

\section{Conclusion}

In this work, we introduced an adaptation of the Octomap framework that fuses external depth and proximity sensor data for probabilistic, volumetric map generation. 
Our results show that fusing both data streams into a cohesive map represents the environment in more detail then either independently.
With respect to future work, we aim to implement this method on a real robot with real sensors, which we have worked with in the past for contact anticipation and avoidance \cite{escobedo2021contact}. Lastly, we plan to expand our simulation to accurately represent the VL53L1X ToF proximity sensor which has a viewing angle of approximately 27\textdegree{} in a cone shape. As objects are sensed further away form a given sensor, the uncertainly in object position increases, because multiple objects of different depth can be sensed from the same sensor position. These specific sensor characteristics must be addressed in order to generate an accurate map of a robot's environment in real life.

\bibliographystyle{IEEEtran}
\bibliography{main}

\end{document}